\documentclass[10pt,a4paper]{article}
\usepackage[left=3cm,right=3cm,top=3cm,bottom=3cm]{geometry}

\usepackage{amsmath}
\usepackage{amsfonts}
\usepackage{amssymb}

\usepackage{subfig}
\usepackage{float}

\usepackage{graphicx} 
\usepackage{booktabs} 
\usepackage{adjustbox}

\usepackage[utf8]{inputenc}
\usepackage{amsmath}
\usepackage{amsfonts}
\usepackage{amssymb}
\usepackage{graphicx}

\usepackage[linesnumbered,boxruled,vlined]{algorithm2e}
\usepackage{hyperref}
\hypersetup{colorlinks=false}
\usepackage{etex} 
\usepackage{arabtex}
\usepackage{scalefnt}
\usepackage{utf8}
\setcode{utf8}
\usepackage{color}
\usepackage{eurosym}
\usepackage{url}
\PassOptionsToPackage{hyphens}{url}
\usepackage{pbox}
\usepackage{multirow}
\usepackage{multicol}
\usepackage{relsize}		
\usepackage{wrapfig}
\usepackage{minibox}
\usepackage{bchart}
\usepackage{wasysym}

\everymath{\displaystyle}
\usepackage{pgfplots}
\pgfplotsset{compat=newest}
\usepackage{tikz,lipsum}
\usetikzlibrary{matrix,decorations.pathreplacing,positioning}
\usetikzlibrary{shapes,shadows,arrows,patterns}
\tikzset{    
    multidocs/.style={double copy shadow={shadow xshift=-0.4ex,shadow yshift=0.4ex},fill=white,draw=black},
}

\SetCommentSty{mycommfont}
\usepackage{xcolor} 
\def\HiLi{\leavevmode\rlap{\hbox to \hsize{\color{yellow!50}\leaders\hrule height .8\baselineskip depth .5ex\hfill}}}

\DeclareFontFamily{U}{xnsh}{}
\DeclareFontShape{U}{xnsh}{m}{n}{                                  
   <-6> sfixed * [3.0] xnsh14
      <6-10> s * [1.2] xnsh14
         <10><10.95><12><14.4><17.28><20.74><24.88> s * [1.2] xnsh14
         }{}
\DeclareFontShape{U}{xnsh}{bx}{n}{
   <-6> sfixed * [3.0] xnsh14bf
   <6-10> s * [1.2] xnsh14bf
   <10><10.95><12><14.4><17.28><20.74><24.88> s * [1.2] xnsh14bf
}{}

\tikzstyle{decision} = [diamond, draw]
\tikzstyle{line} = [draw]
\tikzstyle{arrow} = [draw,>=triangle 45,->]
\tikzstyle{dash} = [draw, dashed]
\tikzstyle{elli}=[ellipse, draw]
\tikzstyle{rec} = [rectangle, draw]
\tikzstyle{decision} = [diamond, draw]
\tikzstyle{round} = [circle, draw]
\tikzstyle{corpus} = [shape=document, multidocs, draw]
\tikzstyle{doc} = [shape=document, draw, minimum width=0.7cm] 

\makeatletter
\pgfdeclareshape{document}{
\inheritsavedanchors[from=rectangle] 
\inheritanchorborder[from=rectangle]
\inheritanchor[from=rectangle]{center}
\inheritanchor[from=rectangle]{north}
\inheritanchor[from=rectangle]{south}
\inheritanchor[from=rectangle]{west}
\inheritanchor[from=rectangle]{east}
\backgroundpath{
\southwest \pgf@xa=\pgf@x \pgf@ya=\pgf@y
\northeast \pgf@xb=\pgf@x \pgf@yb=\pgf@y
\pgf@xc=\pgf@xb \advance\pgf@xc by-10pt 
\pgf@yc=\pgf@yb \advance\pgf@yc by-10pt
\pgfpathmoveto{\pgfpoint{\pgf@xa}{\pgf@ya}}
\pgfpathlineto{\pgfpoint{\pgf@xa}{\pgf@yb}}
\pgfpathlineto{\pgfpoint{\pgf@xc}{\pgf@yb}}
\pgfpathlineto{\pgfpoint{\pgf@xb}{\pgf@yc}}
\pgfpathlineto{\pgfpoint{\pgf@xb}{\pgf@ya}}
\pgfpathclose
\pgfpathmoveto{\pgfpoint{\pgf@xc}{\pgf@yb}}
\pgfpathlineto{\pgfpoint{\pgf@xc}{\pgf@yc}}
\pgfpathlineto{\pgfpoint{\pgf@xb}{\pgf@yc}}
\pgfpathlineto{\pgfpoint{\pgf@xc}{\pgf@yc}}
}
}

\g@addto@macro{\UrlBreaks}{\UrlOrds}
\g@addto@macro{\UrlBreaks}{\do\/\do\a\do\b\do\c\do\d\do\e\do\f\do\g\do\h\do\i\do\j\do\k\do\l\do\m\do\n\do\o\do\p\do\q\do\r\do\s\do\t\do\u\do\v\do\w\do\x\do\y\do\z\do\A\do\B\do\C\do\D\do\E\do\F\do\G\do\H\do\I\do\J\do\K\do\L\do\M\do\N\do\O\do\P\do\Q\do\R\do\S\do\T\do\U\do\V\do\W\do\X\do\Y\do\Z\do\1\do\2\do\3\do\4\do\5\do\6\do\7\do\8\do\9\do\0\do\.}
\makeatother

\usepackage{ulem}

\author{Motaz Saad and David Langlois and Kamel Sma\"{i}li}
\title{Cross-lingual Opinions and Emotions Mining in Comparable Documents}

\newlength{\textlarg}

\begin{document}
\sloppy 
\date{}

\maketitle

\begin{abstract}

Comparable texts are a set of topic aligned documents in multiple languages, which are not necessarily translations of each other. These documents are informative because they can tell what is being said about a topic in different languages. The aim of this research is to study differences in cross-lingual comparable documents in terms of sentiments and emotions. In this work, we focus on English-Arabic language pair. Before comparing sentiments and emotions in comparable documents, texts must be annotated first with sentiment and emotions labels. For that, we use a cross-lingual method to annotate source and target texts with opinion labels (subjective and objective). The advantage of this method is that it produces resources in multiple languages without the need of machine translation system. To be able to annotate the English-Arabic document with emotion labels (anger, disgust, fear, joy, sadness, and surprise), we manually translate the WordNet-Affect (WNA) emotion lexicon from English into Arabic. Then we use these English-Arabic emotion lexicons to annotate English-Arabic comparable documents. Finally, we use a statistical measure to compare the agreement of sentiments and emotions in the source and the target pair of the comparable documents. This comparison is interesting, especially when the source and the target documents come from different sources. To our knowledge, this has not addressed in the literature. Our method inspects the agreement for each pair (source and target) of documents. In our experiment, we study the agreement between English and Arabic document pairs collected from Euronews, British Broadcast Corporation (BBC), and Al-Jazeera (JSC) news websites. The methods presented in this research are language independent and they can be applied to any language pair. Result shows that sentiments and emotions in comparable converge (agree) when the inspected news articles are from the same news agencies, and diverge (disagree) when they are from different news agencies.

\end{abstract}


\section{Introduction}

Retrieving opinions and emotions from comparable multilingual documents can be very useful. For example, a journalist wants to know what is being said around the world about war in Syria. For that, he needs to access and analyze documents about this topic in several languages. This is necessary because the opinions and sentiments expressed in documents depend strongly on the nationality of the source. This is due to cultural differences and geopolitical aspects. Indeed, during the Egyptian revolution in 2011, many non-Arabic speaking journalists were interested in what was being said in Arabic posts in the social media. During that time, automatic translation services were provided by Twitter and other organizations to understand these posts \cite{theguardian,dw}.

\begin{figure}[!tbph]
\begin{center}
\subfloat[The English document]{\fbox{\includegraphics[width=\textwidth,height=\textheight,keepaspectratio]{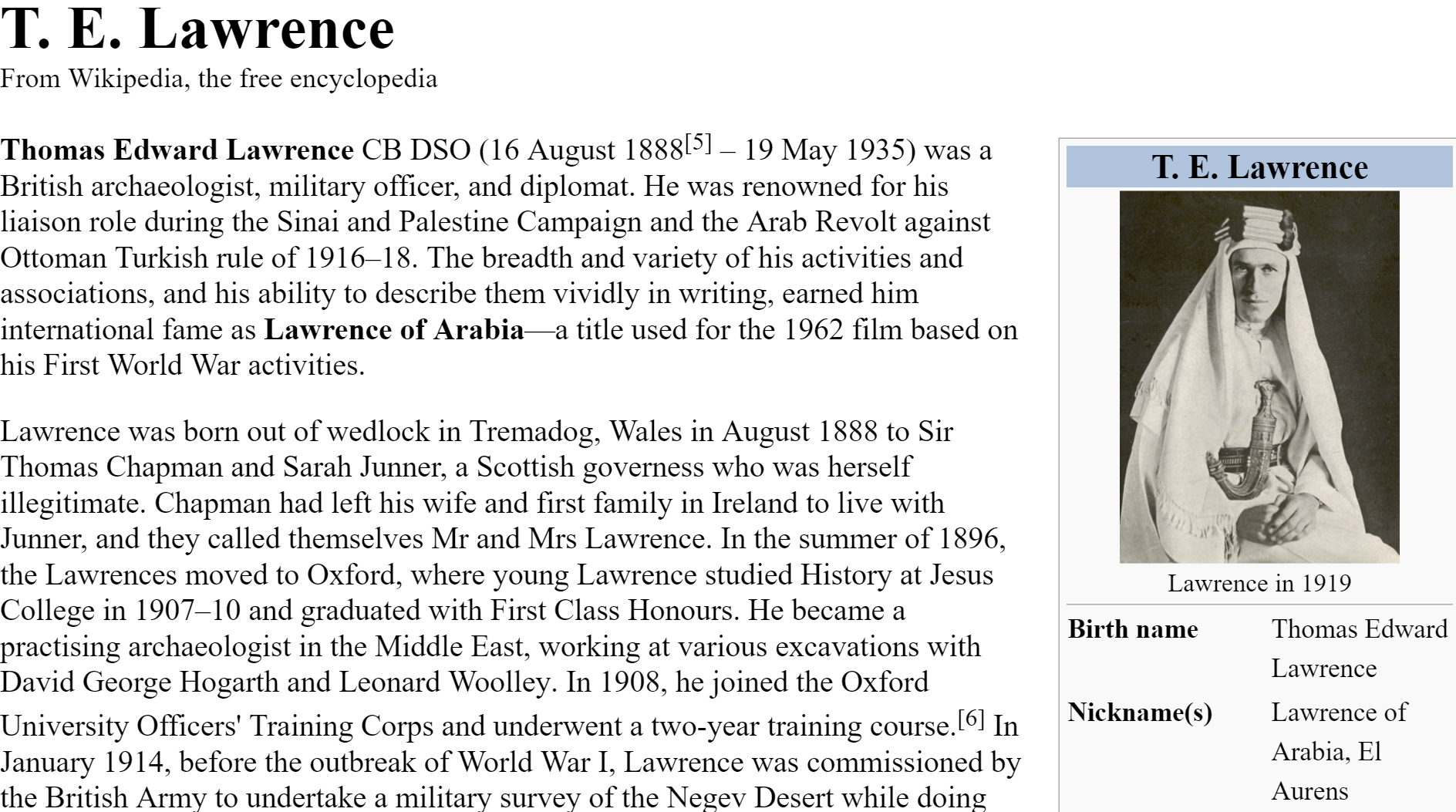}}\label{fig:wiki_en}}
\newline
\subfloat[The French document]{\fbox{\includegraphics[width=\textwidth,height=\textheight,keepaspectratio]{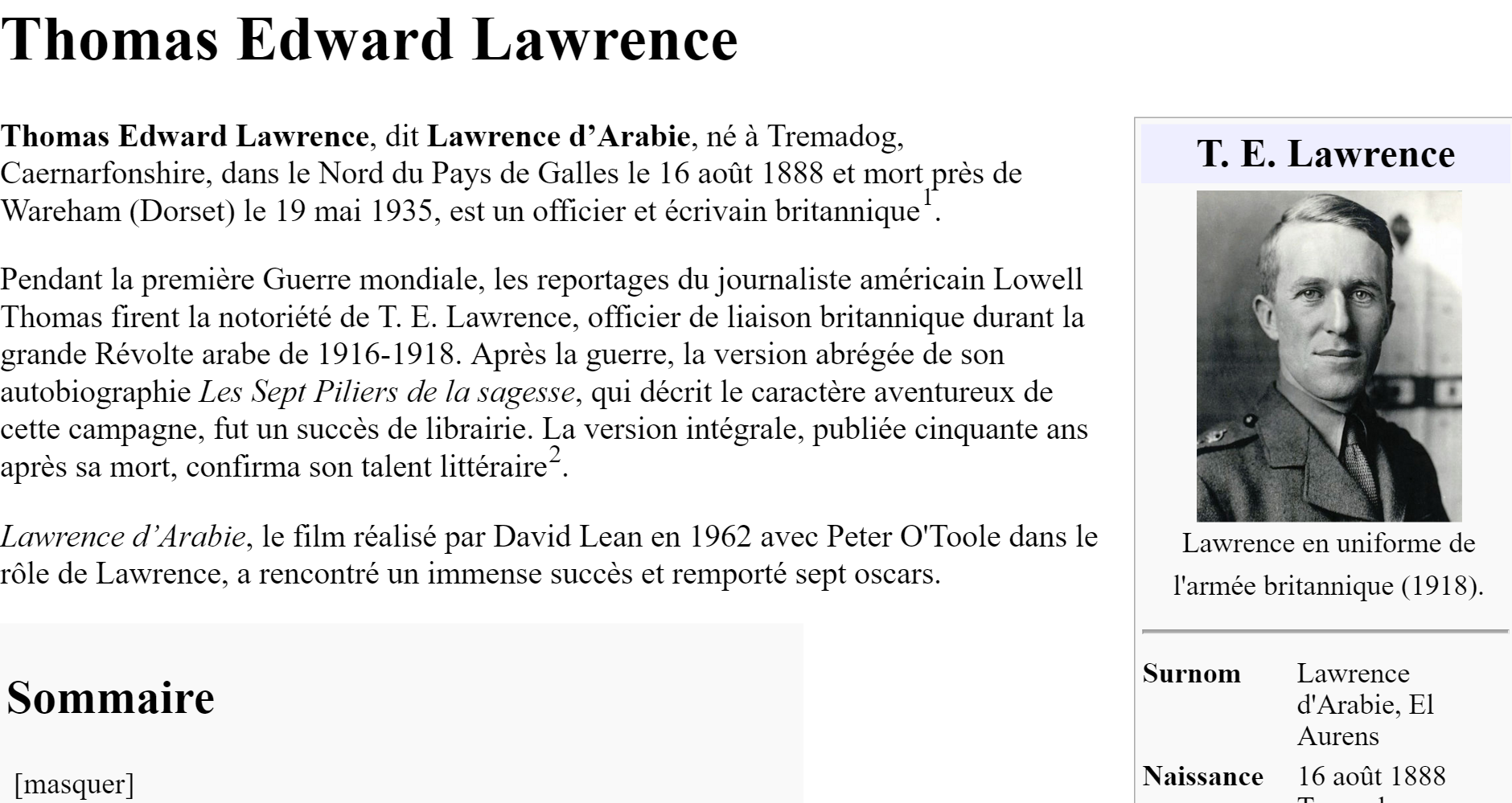}}\label{fig:wiki_fr}}
\end{center}
\caption{English-French comparable documents from Wikipedia}
\label{fig:wikipeida_comparable_lowarence}
\end{figure}

Two documents are called comparable because they contain approximately the same information, i.e., they are related to the same topic. This is the case, for example, for Wikipedia pages about the same entry. Figure \ref{fig:wikipeida_comparable_lowarence} shows an example of English\footnote{\url{https://en.wikipedia.org/wiki/T._E._Lawrence}} and French\footnote{\url{https://fr.wikipedia.org/wiki/Thomas_Edward_Lawrence}} Wikipedia comparable documents related to a biography of a person. It can be noted from the figure that the first paragraph in the English document is longer than the French one, it also provides more information about the person. Even if the documents differ in terms of facts, they remain both objective. 

Another example of comparable documents is the English\footnote{\url{http://www.euronews.com/2013/12/03/transparency-international-warns-23rds-of-states-corrupt/}} and the French\footnote{\url{http://fr.euronews.com/2013/12/03/corruption-l-espagne-recule-dans-l-index-de-transparency-international/}} news articles from Euronews shown in Figure \ref{fig:enews_comparable_corruption}. The news articles are related to a report about transparency and corruption in the world. Despite the documents are related to the same news story, the translations of their titles are different. Furthermore, the first paragraph in the English and in the French documents are different. The French document also has an additional paragraph that describes the position of France in this report. This paragraph does not exist in the English document. Nevertheless the both document remain objective and relate facts.

\begin{figure}[!tbph]
\begin{center}
\subfloat[The English news document]{
\fbox{
\begin{tabular}{p{\columnwidth}}
\textbf{Transparency International warns 2/3rds of states ``corrupt"}
\\
03/12/13 16:59 CET   
\\
The fairtrade watchdog Transparency International reports in its latest global roundup that two-thirds of the 177 countries it surveys are below average in the corruption stakes.
\\
The report highlights continuing political pressure resulting in market distortions caused by bribery, cronyism, a lack of accountability and inadequate legal systems.
\\
``We have two-thirds of all countries, 177 countries in total, where we can see that they score under 50 points, which means they are below the average, and corruption in those countries, two thirds of the 177, are still to be seen in a very critical situation," says Transparency’s German head Edda Muller.
\\
Top of the class is Denmark, followed by New Zealand and Finland. The USA was only 19th, while among the major European economies France, (22), again lagged behind Germany, (12), and the UK, (14).
\\
Greece's position improved, but it is still a lowly 80th, a potential turn-off for investors and the worst in Europe. Somalia, Afghanistan and North Korea tied for last place, unchanged from last year.
\end{tabular}}
\label{fig:enews_en}}
\newline
\subfloat[The French news document]{
\fbox{
\begin{tabular}{p{\columnwidth}}
\textbf{Corruption : l’Espagne recule dans l’index de Transparency International}
\\
03/12/13 16:59 CET  
\\
Selon l’indice de la corruption dans le monde, calculé par l’ONG Transparency International et publié mardi, le Danemark est le pays le plus honnête, à égalité avec la Nouvelle Zélande. Ces deux pays jugé les moins corrompus en 2013, étaient déjà les deux meilleurs de la liste de 177 pays établie par l’ONG basée en Allemagne.
\\
“Nous avons deux tiers des 177 pays qui sont sous les 50 points, explique Edda Muller, la présidente de Transparency International, ce qui veut dire qu’ils sont sous la note moyenne. La corruption dans ces pays : les deux tiers des 177, est dans une situation très critique”.
\\
Les trois premiers du classement sont le Danemark, la Nouvelle Zélande et la Finlande. La Grèce gagne quelques places mais reste 80ème et dernier pays européen.
\\
La surprise c’est l’Espagne en proie aux affaires et qui passe de la trentième à la quarantième place. 
En règle générale les pays du Nord de l’Europe font figure d‘élèves modèles. 
L’enquête de Transparency International n’est pas un classement selon le niveau de corruption mais elle mesure la perception de la corruption en raison du secret qui entoure les pratiques les moins avouables.
\\
La France, 22ème, n’occupe que le 10ème rang en Europe. L’ONG considère que le bilan des lois votées en France en 2013 en matière de transparence et de lutte contre la corruption est “globalement positif” mais s’interroge sur leur mise en oeuvre effective.
\end{tabular}}
\label{fig:enews_fr}}
\end{center}
\caption{English-French comparable news article from EURONEWS}
\label{fig:enews_comparable_corruption}
\end{figure}

But a document can contain subjective paragraphs or sentences. Comparable documents about a political or polemic topic (war, societal topic) may contain very contrasting opinions and sentiments. This can be the case in social medias or blogs. Another example is a review about a product. To illustrate this fact, let's use comments about a holiday home in Avignon (France) posted on a famous commercial web service. For the same house, one have comments in French (3), Italian (2), English (8), Chinese (1), Russian (1) and Dutch (1). In French, we can read positive comments such as \textit{"déco exceptionnelle", "calme", "inestimable", "charmant"}, or negative such as\textit{ "rideaux des chambres ne permettant pas le noir complet."}; in English, we can read \textit{"wonderful", "great house", "perfect", "comfortable", "Private, quiet, nice courtyard and the landlord very accommodating"}, but also negative comments such as \textit{"the website image is better than real. The condition of the building a bit run down. the living room and dinning room is too dim. Got mosquitoes."}. Therefore, here the topic is the same (a review about a house), but there are several languages, and each user gives its own opinion. A future french consumer, not comfortable with English, Italian or Chinese, may want to understand all reviews (because french reviews for this house are not numerous).

Cross-lingual comparable documents have the same topic but are not necessarily translations of each other.  To detect if two comparable documents are objective or subjective is very difficult because natural language is complex and ambiguous. For example, the writer could use implicit or ironic words or expressions. If a document is subjective, another problem arises which is to detect if it is "positive" or "negative". Here also the task is difficult because of natural language and because there may be a mix between general opinion and writer opinion (for example an appeal for sadness sentiment). Last, two comparable documents can express very contrasting sentiments (happy/sadness for a sport result, depending of the nationality of the writer).

The aim of this paper is to describe how sentiments and emotions may differ in comparable documents. To our knowledge, comparing sentiments and emotions in a cross-lingual context has not been addressed in the literature. This research provides and evaluates methods to annotate and to compare the agreement of comparable documents in terms of sentiments. We focus in this work on English-Arabic comparable documents. Studying English-Arabic comparable documents is interesting because sentiments expressed in these documents may diverge for some topics due the political situation in the Arabic regions associated with the emergence of so-called ``Arab Spring". 

To check whether the sentiments or the emotions diverge or converge in the comparable documents, comparable texts must be labeled with sentiment and emotions categories, then one can check the agreement of labeled documents. The process of identifying sentiments in the text is called sentiment analysis or opinion mining \cite{Pang2008}. Sentiment analysis consists in the analysis of the subjectivity or the polarity of the text. Subjectivity analysis classifies a text into subjective or objective, while polarity analysis classifies the text into positive or negative. As for emotions identification, it consists in identifying emotions expressed in the text such as joy, sadness, etc. 

In this work, we need comparable documents to be annotated first with sentiment labels before comparing the agreement of these labels. So we need English-Arabic sentiment resources to build classifiers that can be used to automatically annotate comparable texts. 

The method we use to build these classifiers is based on the hypothesis that for parallel corpora, sentiments expressed in the source part are the same than in the target side. Moreover, Therefore, parallel corpora allow to check if our classifiers manage to label both source and target sentences with same sentiments or emotions category before to apply them onto comparable corpora. Therefore, we need parallel corpora. Parallel texts are made up of a set of aligned sentences which are translations of each other. Parallel corpora are acquired using human translators, but this is time-consuming and requires a lot of human being efforts. Comparable corpora can be obtained easily from the web. The emergence of Web 2.0 technologies enlarged web contents in many languages. Newspaper websites and Encyclopedias are ideal for collecting comparable documents. But aligning these texts is a challenging task. There are many documents in the web which are comparable (i.e., related to the same topic). We describe in Section \ref{sec:corpora} the parallel and comparable corpora we use for this research.

Then, we describe in Section \ref{sec:cross_lingual_annotation} the cross-lingual annotation method \cite {Saad2014lrec} we use to provide these resources. Such kind of annotated resources are useful because they are not available.

Finally, in Section \ref{sec:agreement} we describe the statistical measure we use to compare the agreement of sentiments and emotions in comparable documents. We present experiments in this section on comparable news documents collected from different sources. 

\section{Related Words}
\label{sec:related_work}

Our work focuses on comparing automatic labeling of cross-lingual comparable documents in terms of opinions/sentiments and emotions. For that we need labeled comparable corpora. In section \ref{sec:cross_lingual_annotation}, we describe how we obtain such a resource basing on a previous work \cite{Saad2014lrec}. Consequently, we propose a brief description of related works about sentiment analysis, and afterwards a description about cross-lingual comparison of comparable documents in terms of sentiments and emotions.

\subsection{Sentiment Analysis}

Sentiment analysis (or opinion mining) consists in identifying the subjectivity or the polarity of a text \cite{Pang2008}. Subjectivity analysis includes the classification of a given text into subjective (\textit{e.g., The book is interesting!}) or objective (\textit{e.g., The new edition of the book is released}) labels, while polarity analysis aims to classify the text into positive (\textit{e.g., The image quality is good!}) or negative (\textit{e.g., The battery life is very short!}). 

A subjective (or opinionated) text conveys opinions, while an objective text represents facts. As for polarity, positive and negative texts are opinionated. These opinions can be like/dislike, satisfied/unsatisfied, happy/sad, etc. Opinions can be related to a product, a book, a movie, a news story, etc. Some texts may contain mixed sentiments (\textit{e.g., ``I like this camera, but it is very expensive!"} or \textit{``The phone will be available for order starting from the next week, but the price is prohibitively expensive!"}). Since sentiments labeling is subjective task, it is very difficult task, even humans may disagree among themselves about sentiment labels of a given text. This is why automatically analyzing subjectivity of the texts is a challenging task.

Popular methods for sentiment analysis include lexicon based and corpus based \cite{Pang2008}. The lexicon based method uses a pre-annotated lexicon, which is usually composed of a set of terms and corresponding scores representing the subjectivity or the polarity of the terms. The lexicon based method uses string matching techniques to match words in the text and the terms of the annotated lexicon, then it calculates the average score of the matched words. 

There are many sentiment lexicons such as SentiWordNet \cite{Baccianella2010}, which is an extension of WordNet, and SenticNet \cite{Cambria2010}, which is a knowledge-based lexicon. Both lexicons are built using a semi-automatic method. 

The corpus based method is also a popular approach for sentiment analysis. It uses a pre-annotated corpus and machine learning algorithms to train a classifier to automatically classify given texts. The task is considered as a text classification problem. The most commonly used features include n-grams and POS tags, and the most commonly used classifiers are Support Vector Machines (SVM) and Naive Bayes (NB) \cite{Pang2008}. One of the publicly available resources is a collection of movie reviews, which are pre-annotated with subjectivity labels \cite{Pang2004}, and polarity labels \cite{Pang2005}. We use this resource to bootstrap our sentiment classifiers (see Section \ref{sec:cross_lingual_annotation}).

Most of researchers define multilingual sentiment analysis as generating sentiment resources for low-resourced languages from a high-resourced language, such as English, using Machine Translation technique. There is a debate in the community about whether machine translation is sufficient to capture sentiments or not. To our knowledge, comparing sentiments across languages is only addressed by \cite{Bautin2008}, who also used machine translation to translate all non-English texts into English, then they compared sentiments in these texts. In \cite{Bautin2008}, the authors used machine translation systems to translate texts of eight languages into English, then they applied sentiment analysis on the translated texts. The authors investigated whether machine translation is sufficient to capture sentiments in the translated texts. The authors used a sentiment analysis system called Lydia \cite{Lloyd2006} to analyze sentiments based on named entities. For a given named entity (a person, a city, etc.), the system computes the sentiment score. The score quantifies the polarity of stories related to the named entity in news collected in a time period. The score calculation in Lydia system is done using a pre-defined sentiment lexicon. The authors believed that despite machine translation (MT) makes some serious errors in translations, it can be sufficient to capture sentiments. In the same way, \cite{araujo2016evaluation} conclude that machine translation approach used with English based methods can compete well with specific language methods. \cite{hajmohammadi2015combination} argue that using MT is not so easy because lexical statistics for expressing sentiments are not the same from source and target languages. The source model is therefore sub-optimal for target language. In order to improve the approach,  \cite{hajmohammadi2015combination} propose to translate target data into source language, to detect interesting examples, and to ask to human experts to check the labels in order to feed the model with target data. 

On the other hand, the work of \cite{Brooke2009} explored the adaptation of English resources for sentiment analysis of Spanish texts. The authors examined two approaches: the first one is to build and annotate resources in the Spanish language from scratch, the second one is to use a MT system to translate English sentiment resources into Spanish. The authors compared a sentiment analysis method that uses the translated resources and the resources that they built from scratch. The authors reported that translation has a disruptive effect on the performance of sentiment analysis. Moreover, it is time and effort consuming to translate the lexicon and the corpus. Therefore, the authors concluded that it is worthy to build resources from scratch. 

\subsection{Emotion Identification}

Emotion identification is the automatic detection of emotions that are expressed in a text. It is useful for many applications such as market analysis, text-to-speech synthesis, and human-computer interaction \cite{Pang2008}.

The basic six human emotions, which are reported in a psychological study by \cite{Ekman1992}, are widely adopted in emotion identification \cite{Pang2008}. These emotions are \textit{anger, disgust, fear, joy, sadness,} and \textit{surprise}. 

\cite{Strapparava2007} created the WordNet-Affect (WNA) emotions lexicon by annotating a subset of English WordNet. Each entry (synonym set or synset) of the WNA is annotated with one of the six basic emotions. The WNA can be used as emotion lexicon to detect emotions in texts, or it can be used to extract emotion features. Extracted features can be used to train machine learning algorithms to detect emotions \cite{Alm2005,Aman2007}. In addition, WNA can be also used to develop emotion resources for non-English languages. For instance, \cite{Bobicev2010} translated WordNet-Affect from English into Romanian and Russian languages using a bilingual dictionary. Also \cite{Torii2011} developed a Japanese WNA from the English one by crossing synsets-IDs with the Japanese WordNet. In this work, we translate WNA emotions lexicon from English into Arabic manually in order to use it in our experiments.

\subsection{Comparing sentiments and emotions across languages}

As seen in previous sections, there is a huge effort for building/adapting resources from one language to another. This is natural because to build reference resources for each language is very time consuming.

There are less works on comparing languages in terms of sentiments and emotions. But this could be very useful because opinions about a topic, a news, a product may be very different depending on nationality/culture, and thus depending on language.

The work of \cite{severyn2016multi} is not directly linked to the scope. Indeed, this work is dedicated to automatically label Italian and English youtube videos in terms of opinion (positive/negative/neutral) and target (does the comment focus on the video form, or on the video content). But \cite{severyn2016multi} use a corpus which focuses on two topics: automobiles and tablets. This corpus is analyzed for each language. This analysis gives information on statistics about manual labels. The agreement between human experts is given. But the agreement between languages on the same topic is not described. Our focus in the paper is different because we analyze the cross-lingual agreement for two news about the same topic.

\cite{gutierrez2016detecting} propose to analyze multi-lingual corpora in terms of cultural differences. They use a LDA approach in order to automatically detect topics, topic-words and perspective-words. Topic-words indicate words related to the neutral description of a topic while perspective-words are related to the way the writer handle the topic. The authors show lexical differences on the same topics between English, Spanish and Russian. The level of analysis is the corpora: in general, how Spanish people speak about a specific topic? Differently, we focus on document level: we study a couple of bi-lingual documents about the same topic and we explore the agreement of automatic labeling in terms of subjectivity/objectivity and emotions.


\section{The Used Corpora}
\label{sec:corpora}

This section presents the parallel and comparable corpora used in our work. We describe below three corpora, one is parallel, two are comparable. We will study the cross-lingual agreement of these corpora in Section \ref{sec:agreement}. 

\subsection{The Parallel Corpora}

We use parallel corpora to produce annotated resources using the cross-lingual projection method. The parallel corpus is also used as a baseline of sentiment agreement in comparable documents. Indeed, two aligned sentences are translation from each others. Therefore, they contain exactly the same information and the same content in terms of sentiment and opinion. This can be useful to check if a method can detect agreement across languages.

Table~\ref{tab:corpus_parallel} shows the characteristics of the parallel corpora that we use in this work. $\vert S \vert$ is the number of sentences, $\vert W \vert$ is the number of words, and $\vert V \vert$ is the vocabulary size. The table also shows the domain of each corpus. These corpora come from several different domains, and they are ideal to test our methods on different genres of texts. These corpora include 
AFP\footnote{\url{www.afp.com}}, ANN\footnote{\url{www.annahar.com}}, ASB\footnote{\url{www.assabah.com.tn}} \cite{Dalal2009}, Medar\footnote{\url{www.medar.info}}, NIST  \cite{NIST}, UN \cite{Rafalovitch2009}, and TED\footnote{\url{www.ted.com}} \cite{Cettolo2012}. The corpora are collected from different sources and present different genres of text. It is remarkable that AFP, ANN, ASB, Medar, NIST and UN corpora are generated by professional translators, while TED are generated by volunteer translators

\begin{table}[!htb]
\caption{Parallel news Corpora characteristics}
\label{tab:corpus_parallel}
\centering
\begin{tabular}{|c|c|c|c|c|c|}
\hline 
\multirow{2}{*}{\bf Corpus} & \multirow{2}{*}{$\vert S \vert$} & \multicolumn{2}{|c|}{$\vert W \vert$} & \multicolumn{2}{|c|}{$\vert V \vert$}  \\ 
  &   &  English & Arabic &  English & Arabic \\
\hline 
\multicolumn{6}{|l|}{\bf Newspapers}    \\    
\hline          
AFP &   4K & 140K & 114K & 17K & 25K  \\ 
ANN &   10K & 387K & 288K & 39K & 63K \\ 
ASB &   4K & 187K & 139K & 21K & 34K \\ 
Medar &  13K & 398K & 382K & 43K & 71K \\ 
NIST &   2K & 85K & 64K & 15K & 22K \\ 
\hline
\multicolumn{6}{|l|}{\bf United Nations Resolutions}    \\              
\hline
UN &    61K & 2.8M & 2.4M & 42K & 77K \\ 
\hline
\multicolumn{6}{|l|}{\bf Talks}    \\   
\hline           
TED &   88K & 1.9M & 1.6M & 88K & 182K \\ 
\hline
\end{tabular} 
\end{table}

We merge parallel news corpora (AFP, ANN, ASB, Medar and NIST) into one corpus, and we call it \textit{parallel-news}. We merge these corpora because they belongs to the same domain (newspapers). The characteristics of this corpus are presented in Table \ref{tab:parallel_news}. 

\begin{table}[!htb]
\caption{\textit{parallel-news} corpus characteristics}
\label{tab:parallel_news}
\centering
\begin{tabular}{|c|c|c|c|c|c|}
\hline 
 \multirow{2}{*}{$\vert S \vert$} & \multicolumn{2}{|c|}{$\vert W \vert$} & \multicolumn{2}{|c|}{$\vert V \vert$}  \\ 
 & English & Arabic &  English & Arabic \\
\hline           
34K & 1.2M & 0.9M & 83K & 141K  \\ 
\hline 
\end{tabular} 
\end{table}

\subsection{The Comparable Corpora}

We use comparable corpora to study sentiments and emotions agreement in comparable documents. The objective is to investigate sentiments and emotions agreement of news articles that come from the Arab media with the ones that come from the English media. The used corpora include Euronews and BBC-JSC. Euronews\footnote{\url{www.euronews.com}} is a set of aligned news articles collected and aligned by \cite{Saad2013cilc}, and BBC-JSC corpus is a set of aligned news articles collected and aligned by \cite{Saad2015phd} from British Broadcast Corporation\footnote{\url{www.bbc.co.uk}} (BBC) and Al-Jazeera\footnote{\url{www.aljazeera.net}} (JSC).


Euronews corpus contains about 34K comparable articles as shown in Table \ref{tab:euronews_corpus}. $\vert D \vert$ in the table is the number of documents in the corpus, $\vert W \vert$ is the number of words in the corpus, and $\vert V \vert$ is the vocabulary size. The BBC-JSC corpus is composed of 305 aligned news articles as shown in Table \ref{tab:bbcjsc_corpus}.

\begin{table}[!htb]
\caption{Euronews comparable corpus characteristics}
\label{tab:euronews_corpus}
\centering
\begin{tabular}{|c|c|c|c|}
\hline 
 & English &  Arabic \\
\hline 
$\vert D \vert$  & 34K & 34K \\
$\vert W \vert$ &  6.8M &  5.5M \\
$\vert V \vert$ &  232K & 373K \\
\hline 
\end{tabular}
\end{table}

\begin{table}[!htb]
\caption{BBC-JSC comparable corpus characteristics}
\label{tab:bbcjsc_corpus}
\centering
\begin{tabular}{|c|c|c|c|}
\hline 
 & English &  Arabic \\
\hline 
$\vert D \vert$  & 305 & 305 \\
$\vert W \vert$ &  75K &  45K \\
$\vert V \vert$ &  9K & 12K \\
\hline 
\end{tabular}
\end{table}

\section{Cross-lingual Sentiment Annotation}
\label{sec:cross_lingual_annotation}

Our work focuses on comparing the cross-lingual agreement between two parallel or comparable documents in terms of sentiments and emotions. For that we need to label each document with positive/negative class by using a classifier. 

We used the classifiers built in a previous work \cite{Saad2014lrec}. In sake of clarity, we give below a brief overview of these classifiers. 

Most of state-of-the-art focus only on creating sentiment resources for low-resourced languages by building these resources from scratch or by adapting English resources using machine translation systems. On the other hand, many authors have argued whether machine translation preserves sentiments \cite{Denecke2008,Ghorbel2012}. Our method for creating sentiment resources in multiple languages is based on the hypothesis that a sentiment annotation model trained for a domain can be used for another domain, and that sentiment annotation can be transferred from one language to another by using parallel corpus. We use this method to create resources in English and Arabic languages in various domains. Next, we use these resources to build sentiment classifiers which can be used to automatically annotate English-Arabic articles. The advantage of this method is that it does not require a machine translation system. Sentiment resources in English-Arabic languages are not available; therefore, creating these resources is one of the contributions of this research.

In order to build a classifier for English and for Arabic, we start with a classifier for English, we use this classifier to label the English part of an English-Arabic parallel corpus, we report the obtained tags from the English part to the Arabic part. Then we estimate the Arabic classifier basing on this Arabic labeled corpus.

The initial English classifier is trained on a collection of movie reviews written in English \cite{Pang2004}. This corpus is pre-annotated with subjective and objective labels. The corpus is composed of 5,000 subjective and 5,000 objective sentences. The authors collected the subjective reviews from the Rotten Tomatoes website\footnote{www.rottentomatoes.com}, and the objective reviews from IMDb plot summaries\footnote{www.imdb.com}. Rotten Tomatoes website is launched in 1998 and it is dedicated to film reviews. The website is widely known as a worldwide film review aggregator for important critics. The website also enable users to review and discuss films. IMDb stands for Internet Movie Database, and has been launched in 1990. It is on-line database of information related to films, TV programs, actors, plot summaries, etc. A plot summary tells the main things that happened in the film. It is a brief description of the story of the film. 

The classifier is based on the Naive Bayes approach where each document is described by binary features: occurrence of 1-grams, 2-grams and 3-grams. Only n-grams occurring twice are more in training corpus are kept as features.

In \cite{Saad2014lrec}, we evaluated the obtained classifiers on a subset of the parallel corpora described in Section \ref{sec:corpora}. The classifiers obtain accuracy equal to 0.718 on parallel news corpora, and 0.658 on parallel non-news corpora. Moreover, we automatically annotated each parallel corpus of section \ref{sec:corpora} with subjective and objective labels. The obtained labels agree with the nature of the corpora: documents from newspaper are in majority labeled as "objective", while corpus from TED talks are labeled in majority as "subjective".

Basing on these results, we use the English and Arabic classifiers in order to automatically label BBC-JSC and Euronews comparable corpus in terms of objectivity/subjectivity and we study the agreement between classifiers in next section.

\section{Comparing Sentiments and Emotions in Comparable Documents}
\label{sec:agreement}

As outlined in section \ref{sec:related_work}, for multilingual sentiment analysis, the major interest is to create sentiment resources for low-resourced languages, or to demonstrate whether machine translation is sufficient to capture sentiments. To our knowledge, comparing sentiments or emotions in comparable texts is not addressed in the literature. Comparing sentiments or emotions in comparable texts consists in inspecting the agreement of the expressed sentiments or emotions in the source and the target texts. The new contribution in this section is that we compare comparable documents in terms of sentiments and emotions.

For a given pair of English-Arabic documents annotated with sentiments and emotions, one can ask the following question: do these documents convey the same sentiments or emotions? The answer can be provided by a measure of the agreement of the annotations of English and Arabic comparable texts. 

In this section, we study the agreement of annotations in three types of news corpora: a parallel news corpus, and two comparable corpora (Euronews and BBC-JSC). In the next sections, we first describe the used agreement measures. Then we present the method and the experimental results of comparing news documents in terms of sentiments and emotions.

\subsection{Agreement Measures}
\label{sec:agreement_measures}

The measure which we use in this work is called Inter-annotator agreement. It is defined as the degree of agreement or homogeneity between annotators \cite{Artstein2008}. The terms inter-annotator, inter-rater, and inter-coder are used interchangeably in the literature. 

Normally, inter-annotator agreement is used in machine learning when there is no validation set to evaluate the accuracy of the classifier. Inter-annotator agreement is based on the following assumption: if annotations are consistent, then annotators implicitly have similar understanding of the annotation guidelines, which describe how to make the annotations. Consequently, the annotation scheme is expected to perform consistently under these guidelines. In other words, the annotation scheme is reliable if the annotations are consistent (agree to each others). 

In our work, the motivation for studying the agreement of comparable news documents is that if two persons disagree with each others when describing a news event, then it is difficult for a third person to understand what really happened, or maybe the third person thinks that both sources are interesting for him/her because they carry different view points. The idea is to automatically detect the agreement or disagreement between news stories written by different news agencies in different languages. The inspection of agreement of sentiments and emotions in comparable news documents can be seen as follows. Two news agencies $A$ and $B$ try to cover a news event. Each agency writes a document to describe and/or comment on this event. If $A$ and $B$ have the same perspective on the news event, then they will have some degree of agreement in their documents. Otherwise, they will diverge in terms of sentiments and emotions that are expressed in their documents. 

We use inter-annotator agreement to inspect the agreement between sentiments or emotions in the annotated English and Arabic comparable documents. The parallel news corpus is expected to have a ``perfect agreement". If near perfect agreement is achieved for parallel corpus, one can claim that the model is reliable, and it can be used to inspect the agreement between comparable documents. In our work, we develop an annotation scheme (a classifier), we prove that it is reliable for the parallel texts, and then we use this annotation scheme (the classifier) to inspect the agreement between the comparable texts.

Inter-annotator agreement can be calculated using statistical measures, such as Cohen's Kappa ($k$) \cite{Cohen1960}. Unlike the simple percentage agreement calculation, statistical measures take into account the agreement that occurred by chance. Cohen's Kappa ($k$) can be used for the cases where two annotators classify the data into two categories \cite{Artstein2008}, and it is calculated as follows: 

\begin{equation}
\label{eq:kappa}
k = \frac{A_o-A_e}{1-A_e} 
\end{equation} 

where $A_o$ is the observed agreement, and $A_e$ is the expected agreement occurred by chance. $A_e$ is calculated as follows:

\begin{equation}
\label{eq:A_e}
A_e = p(l_1|c_1) \times p(l_2|c_1) + p(l_1|c_2) \times p(l_2|c_2)
\end{equation} 

where $p(l_i|c_j)$ is the probability that the coder $j$ annotates data with the label $i$.

\begin{figure}[!htb]
\centering
\begin{tikzpicture}
\draw (-2,2) -- (10,2);
\draw (-2,1) -- (10,1);
\draw (-2,0) -- (10,0);

\foreach \Point in {(0,2), (2,2), (4,2), (6,2), (8,2), (10,2)}
{ \node at \Point {$\bullet$}; }

\foreach \Point in {(6.7,1), (8,1), (10,1)}
{ \node at \Point {$\bullet$}; }

\foreach \Point in {(4,0), (7.5,0), (10,0)}
{ \node at \Point {$\bullet$}; }

\draw (0,2) node[anchor=south] {0.0};
\draw (2,2) node[anchor=south] {0.2};
\draw (4,2) node[anchor=south] {0.4};
\draw (6,2) node[anchor=south] {0.6};
\draw (8,2) node[anchor=south] {0.8};
\draw (10,2) node[anchor=south] {1.0};
\draw (6.7,1) node[anchor=south] {0.67};
\draw (8,1) node[anchor=south] {0.8};
\draw (10,1) node[anchor=south] {1.0};
\draw (4,0) node[anchor=south] {0.4};
\draw (7.5,0) node[anchor=south] {0.75};
\draw (10,0) node[anchor=south] {1.0};

\draw (-2,2) node[anchor=south, font=\smaller] {\cite{Landis1977}};
\draw (-2,1) node[anchor=south, font=\smaller] {\cite{Krippendorff1980}};
\draw (-1,0) node[anchor=south, font=\smaller] {\cite{Green1997,Fleiss2013}};

\draw (-1,2) node[anchor=north, font=\tiny] {no agreement};
\draw (1,2) node[anchor=north, font=\footnotesize] {slight};
\draw (3,2) node[anchor=north, font=\footnotesize] {fair};
\draw (5,2) node[anchor=north, font=\footnotesize] {moderate};
\draw (7,2) node[anchor=north, font=\footnotesize] {substantial};
\draw (9,2) node[anchor=north, font=\footnotesize] {perfect};

\draw (3,1) node[anchor=north, font=\footnotesize] {discard};
\draw (7.5,1) node[anchor=north, font=\footnotesize] {tentative};
\draw (9,1) node[anchor=north, font=\footnotesize] {good};

\draw (1,0) node[anchor=north, font=\footnotesize] {low/poor};
\draw (6,0) node[anchor=north, font=\footnotesize] {fair/good};
\draw (8.8,0) node[anchor=north, font=\footnotesize] {high/excellent};

\end{tikzpicture}
\caption{Kappa interpretation}
\label{fig:agreement_scale}
\end{figure}
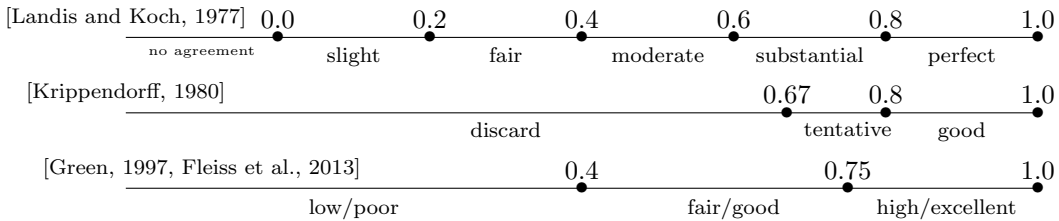

The range of $k$ is between -1 and 1. Various researchers have interpreted the value of $k$ measure in different ways as shown in Figure \ref{fig:agreement_scale}. For example, in \cite{Landis1977}, the authors divided the scale into detailed intervals, where the perfect agreement is assumed if $k>$ 0.8, 0.6-0.8 as substantial agreement, 0.4-0.6 as moderate, 0.2-0.4 as fair, 0.0-0.2 as slight, and $<$ 0 as no agreement. In \cite{Krippendorff1980} the authors discard the agreement if the $k$ value is below 0.67. The $k$ value between 0.67 and 0.8 is considered as tentative agreement, while $k$ above 0.8 means good agreement. In \cite{Green1997,Fleiss2013} the authors consider that $k$ value that is below 0.4 is low/poor agreement, from 0.4 to 0.75 is fair/good agreement, and above 0.75 is high/excellent agreement.

The interpretation of the scale depends on the task. Normally, the agreement for objective annotations is higher than for the subjective annotations \cite{Fort2011}. Examples of objective annotation tasks are POS tagging, syntactic annotation, and phonetic transcription.  Examples of subjective annotation tasks are lexical semantic (subjective interpretation), discourse annotation, and subjectivity analysis. According to \cite{Fort2011}, the agreement for objective tasks can be in the range 0.93-0.95, while for subjective tasks it is in the range 0.67-0.70.

In this work, comparing sentiments and emotions in comparable documents is done as follows:

\begin{enumerate}
\item Automatically annotate comparable documents with sentiment and emotion labels.
\item Inspect the agreement between labels of comparable documents using $k$  agreement measure. 
\end{enumerate}

In the next sections, we describe in detail how automatic annotation is done for comparable documents, then we present the experimental results. Experiments are done on a parallel news corpus, and two comparable corpora (Euronews and BBC-JSC). The parallel-news and Euronews corpora are described in Section \ref{sec:corpora}. We take a subset (10\%) from parallel-news and Euronews to achieve our experiment. Thus, parallel-news is composed of 3.4K sentences and Euronews corpus is composed of 4.2K comparable. Regarding  BBC-JSC corpus, it is composed of 305 comparable documents (see Section \ref{sec:corpora}). In all corpora, each source text of these corpora is aligned to the target one. First we make annotation for the source and the target texts, and then we compare their annotations. The annotation is made at the sentence level for parallel-news corpus, and at the document level for Euronews and BBC-JSC corpora.

\subsection{Comparing Sentiments in Comparable Documents}
\label{sec:agreement_opinions}

The automatic sentiments annotation is done using the classifiers, described in Section \ref{sec:cross_lingual_annotation}. Each pair of comparable documents is annotated with subjective and objective labels using these classifiers.  

\begin{figure}[!htb]
\centering
\begin{tikzpicture}
  \begin{axis}[
    ybar, ymin=0,
    enlarge y limits={upper, value=0.1},
    legend style={at={(0.5,-0.2)},
      anchor=north,legend columns=-1},
    ylabel={Agreement (Kappa)},
    xlabel={Corpus},
    symbolic x coords={BBC-JSC,Euronews,Parallel-news},
    xtick=data,
    grid=major,
    nodes near coords, 
	every node near coord/.append style={font=\tiny,/pgf/number format/fixed, /pgf/number format/precision=2},
	nodes near coords align={vertical},
    ]
    \addplot coordinates {(BBC-JSC,0.06) (Euronews,0.29) (Parallel-news,0.76)};
  \end{axis}
\end{tikzpicture}
\caption{Subjective and objective agreement of news documents}
\label{fig:cross_lingal_subjective}
\end{figure}
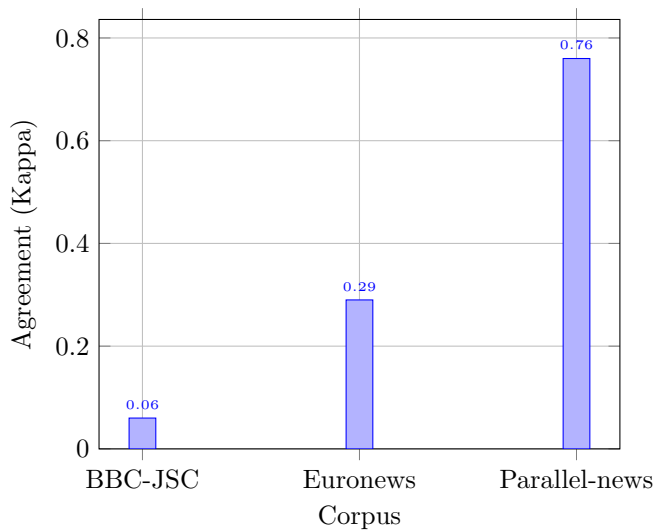

Figure \ref{fig:cross_lingal_subjective} shows the average pairwise subjective and objective agreement (Kappa) calculated for each pair of documents from parallel-news, Euronews and BBC-JSC corpora.

It can be noted from Figure \ref{fig:cross_lingal_subjective} that, as expected, the agreement between source and target texts of the parallel-news corpus can be considered as nearly perfect/good or high according to the interpretations presented in Figure \ref{fig:agreement_scale}. The parallel-news corpus also has the highest agreement scores among other corpora. The Euronews corpus comes in the second place in terms of agreements, and  BBC-JSC corpus has the lowest agreement among the other corpora. For BBC-JSC corpus, the results reveal that BBC documents diverge from JSC documents in terms of subjectivity. These results also show that Euronews documents have higher degree of agreement compared to BBC-JSC. This result is in agreement with the fact that  Euronews documents are mostly translations of each other as mentioned in Section \ref{sec:corpora}, and they are written by the same news agency, while BBC-JSC documents are written by different news agencies. We also used Krippendorff’s alpha \cite{Krippendorff1980} agreement measure on the same corpora, and we had the same conclusions. See \cite{Saad2015phd} for more details.

\subsection{Comparing Emotions in Comparable Documents}
\label{sec:agreement_emotions} 

To identify emotions in comparable documents, we use WordNet-Affect (WNA) emotion lexicon \cite{Strapparava2007}, which is a subset of English WordNet. Each entry (synset) in this lexicon is annotated with one of six emotions (anger, disgust, fear, joy, sadness, and surprise), which are considered as the basic human emotions according to the psychological study conducted in  \cite{Ekman1992}.

To be able to use this lexicon on Arabic texts, we manually translated it into Arabic. Table \ref{tab:wne} describes the English and the Arabic lexicons, where $\vert syn \vert$ is the number of synsets (synonym words are grouped into sets and called synsets), and $\vert w \vert$ is the number of words associated with each emotion label. We use these lexicons to identify emotions in English-Arabic comparable documents.

\begin{table}[!htb]
\caption{English-Arabic WordNet-Affect emotions lexicon}
\label{tab:wne}
\centering
\begin{tabular}{|l|r|r|r|}
\hline 
 Emotion & $\vert syn \vert$ & English $\vert w \vert$ & Arabic $\vert w \vert$ \\ 
\hline  
anger    & 127 &  351   &  748    \\ 
disgust  & 19  &   83   &   155   \\ 
fear     & 82  &   221  &   425   \\ 
joy      & 227 &  543   &    1156  \\  
sadness  & 123 &   259  &   522    \\ 
surprise & 28  &  94    &   201   \\ 
\hline  
Total    & 606  &  1551  &  3207    \\ 
\hline 
\end{tabular} 
\end{table}

To identify whether an emotion is expressed in a text or not using WNA lexicon, the text is first converted into a list of bag-of-words (BOW). To improve the matching between the BOW of the text and words in the lexicons, lemmatization \cite{Miller1998} for English texts and light stemming \cite{Larkey2007Light,Saad2010morphological} for Arabic texts are applied. Each term in the BOW list is checked with the emotion lexicon. If the term is matched with an emotion word in the lexicon, then the emotion label of that word is extracted from the lexicon, and the text is annotated with that label. In the end, the text is associated with six labels which indicate the presence or the absence of emotions that are expressed in the text.

Before using this lexicon to achieve our objective, we need to investigate its performance for identifying emotions. For this purpose, we select a 100 random sentences from news-parallel corpus. Each sentence is annotated with emotion labels by a human reader. The human annotator reads each sentence and check the emotions that are expressed in text, then annotates the text with the corresponding emotion labels. For instance, for the sentence \textit{``Shock and deep sadness in the country due to the sudden death of President"}, then the human annotator annotates this sentence with \textit{surprise} and \textit{sadness} labels. These sentences are also annotated automatically using the WNA lexicon as described above. The automatic annotation is then compared to the human annotation. The evaluation is presented in Table \ref{tab:lexicon_accuracy}. The table shows the accuracy, precision ($P$), recall ($R$), and the F-Measure ($F1$). The accuracy of emotion identification using WNA lexicon ranges between 0.85 and 1.0, while the F-Measure ($F1$) ranges between 0.81 and 1.0. As can be seen in the results, the WNA lexicon can be reliable for emotion identification task. 

\begin{table}[!htb]
\caption{Evaluating WNA emotion lexicon}
\label{tab:lexicon_accuracy}
\centering
\begin{tabular}{|l|c|c|c|c|}
\hline 
Emotion & accuracy & $P$ & $R$ & $F1$ \\
\hline
anger     & 0.91  & 0.95  &  0.70   & 0.81  \\ 
disgust   & 0.98  & 1.00  &  0.75   & 0.86  \\ 
fear      & 0.97  & 1.00  &  0.80   & 0.89  \\ 
joy       & 0.85  & 0.85  &  0.79   & 0.82  \\ 
sadness   & 0.98  & 0.86  &  1.00   & 0.92  \\ 
surprise  & 1.00  & 1.00  &  1.00   & 1.00 \\ 
\hline 
\end{tabular} 
\end{table}

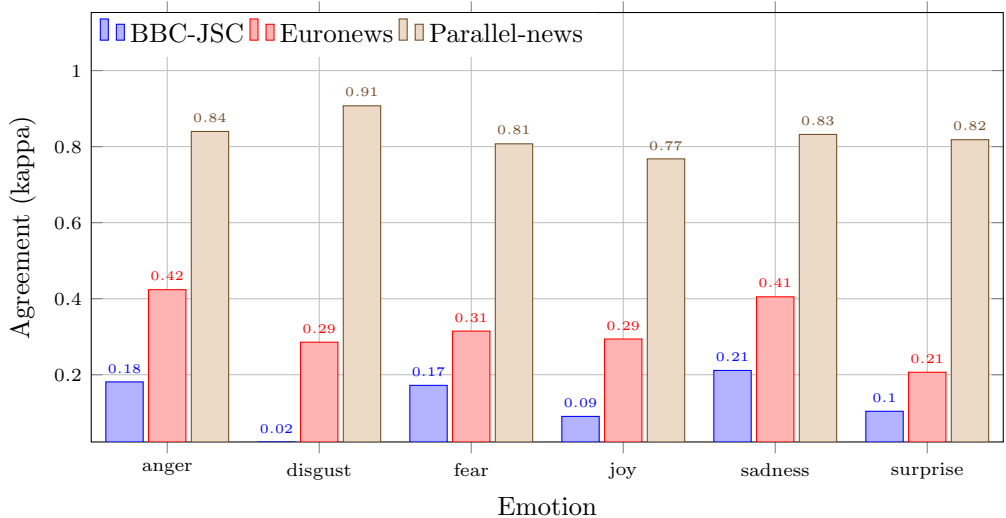
\begin{figure}[!htb]
\centering 
\begin{tikzpicture}
\begin{axis}
[
	ybar, ymax=1.05,
	enlarge y limits={upper, value=0.1},
	legend style={at={(0,1)},anchor=north west,draw=none, legend columns=0},
	ylabel={Agreement (kappa)},
	xlabel={Emotion},
    symbolic x coords={anger, disgust, fear, joy, sadness, surprise},
    tick label style={font=\scriptsize},
    xtick=data,
    x=2cm,
    grid=major,
    bar width=14pt,
    nodes near coords, 
	every node near coord/.append style={font=\tiny,/pgf/number format/fixed, /pgf/number format/precision=2},
	nodes near coords align={vertical}, 
]

\addplot coordinates {
(anger, 0.1812)
(disgust, 0.0232)
(fear, 0.1721)
(joy, 0.0904)
(sadness, 0.2113)
(surprise, 0.1039)

};

\addplot coordinates {
(anger, 0.4237)
(disgust, 0.2856)
(fear, 0.3148)
(joy, 0.2940)
(sadness, 0.4050)
(surprise, 0.2066)

};

\addplot coordinates {
(anger, 0.8399)
(disgust, 0.9074)
(fear, 0.8074)
(joy, 0.7678)
(sadness, 0.8321)
(surprise, 0.8182)

};

\legend{BBC-JSC, Euronews , Parallel-news}    
\end{axis}
\end{tikzpicture}
\caption{Average pairwise agreement of each emotion category in the English-Arabic news documents}
\label{fig:cross_lingal_emotions}
\end{figure}

Using the WNA English and Arabic emotion lexicons, we automatically annotate each pair of documents in parallel-news, Euronews and BBC-JSC corpora. Then, the pairwise label agreement  is calculated for each pair of documents. Figure \ref{fig:cross_lingal_emotions} shows the average pairwise agreement between each emotion category in English-Arabic documents. As can be seen from the results, the parallel-news corpus has the highest agreement score among the other corpora for all emotions. This is expected since this corpus is parallel. In addition, the agreement degree can be considered as good or near perfect for parallel-news according to the scale interpretation that is described in Figure \ref{fig:agreement_scale}.

As shown in Figure \ref{fig:cross_lingal_emotions}, agreement scores for Euronews are higher than for BBC-JSC for all emotions. This shows that emotions expressed in BBC diverge from the ones in JSC for the same news stories in our dataset. This may be because Euronews documents come from the same agency, while BBC-JSC documents are written by different agencies. We also used Krippendorff’s alpha \cite{Krippendorff1980} agreement measure on the same corpora, and we had the same conclusions. See \cite{Saad2015phd} for more details.

In this section, we compared English-Arabic comparable news documents based on sentiments and emotions. The comparison is done in two steps: annotate comparable documents automatically, and inspect the agreement of annotations between annotated documents. 

We provided in this section a method to study comparable documents by comparing sentiments and emotions in these documents and inspect the agreement between sentiments and emotions in the comparable documents.

Results according to agreement scale demonstrated that parallel news documents have high agreement scores, while Euronews documents have some degree of agreement, and BBC-JSC documents have the lowest agreement among other corpora. The results revealed that, for our collected corpora, news documents coming from the same news agency have a higher agreement degree compared to the documents from different news agencies. That is, they diverge from each others in terms of sentiment and emotion agreement. 

\section{Conclusion}

In this paper, we study sentiments and emotions in a cross-lingual context. We describe a cross-lingual projection method to project sentiment labels from one language to another. We also introduce a method to inspect the agreement of sentiments and emotions in comparable documents.

Our proposed cross-lingual projection method is used to build  English-Arabic corpus annotated with subjective and objective labels. The method projects sentiment annotations from one topic domain to another and from one language to another. The advantage of this method is that it produces the annotated resources in multiple languages without need for a machine translation system. We used the resulting corpus to build sentiment classifiers (one for each language). We use these classifiers to annotate comparable documents with sentiment labels.

Finally, we compare in this paper sentiments and emotions in comparable documents. The results are interesting, especially when the source and the target documents come from different sources. The comparison is done by inspecting the pairwise agreement of sentiments and emotions expressed in the source and the target comparable documents using statistical agreement measures. We study the agreement of sentiments in comparable news corpora (Euronews and BBC-JSC). The experiments show that BBC-JSC documents diverge from each other in terms of sentiments and emotions, while the Euronews corpus has a higher agreement, and most of the parallel-news corpus documents express the same sentiments and emotions. Studying comparable documents is a promising research field. The contribution in this research is that we provided language independent methods to study comparable documents in different aspects. We a method to compare sentiments and emotions in comparable document pairs using statistical agreement measures.

In future, we will collect and study comparable documents collected from other sources and in other languages. In addition, more advanced methods for sentiment and emotion annotation will be developed. Recall that we used Naive Bayes classifiers trained on 3-gram features for sentiment annotation, and lexicon based method for emotion annotation. Domain adaptation methods will be used in the future to adapt domains in our cross-lingual annotation method. Other methods for statistical agreement, such as Point-wise Mutual Information (PMI), will be used in the future to compare the agreement of sentiments and emotions between comparable documents.


\medskip
\bibliography{myRef}

\begin{thebibliography}{}

\bibitem[Alm et~al., 2005]{Alm2005}
Alm, C.~O., Roth, D., and Sproat, R. (2005).
\newblock Emotions from text: Machine learning for text-based emotion
  prediction.
\newblock In {\em Proceedings of the Conference on Human Language Technology
  and Empirical Methods in Natural Language Processing}, HLT '05, pages
  579--586, Stroudsburg, PA, USA. Association for Computational Linguistics.

\bibitem[Aman and Szpakowicz, 2007]{Aman2007}
Aman, S. and Szpakowicz, S. (2007).
\newblock Identifying expressions of emotion in text.
\newblock In Matoušek, V. and Mautner, P., editors, {\em Text, Speech and
  Dialogue}, volume 4629 of {\em Lecture Notes in Computer Science}, pages
  196--205. Springer Berlin Heidelberg.

\bibitem[Araujo et~al., 2016]{araujo2016evaluation}
Araujo, M., Reis, J., Pereira, A., and Benevenuto, F. (2016).
\newblock An evaluation of machine translation for multilingual sentence-level
  sentiment analysis.
\newblock In {\em Proceedings of the 31st Annual ACM Symposium on Applied
  Computing}, SAC '16, page 1140–1145, New York, NY, USA. Association for
  Computing Machinery.

\bibitem[Arthur, 2011]{theguardian}
Arthur, C. (2011).
\newblock Google and twitter launch service enabling egyptians to tweet by
  phone.
\newblock
  \url{http://www.theguardian.com/technology/2011/feb/01/google-twitter-egypt},
  [Online; accessed 21-August-2014].

\bibitem[Artstein and Poesio, 2008]{Artstein2008}
Artstein, R. and Poesio, M. (2008).
\newblock Inter-coder agreement for computational linguistics.
\newblock {\em Computational Linguistics}, 34(4):555--596.

\bibitem[Baccianella et~al., 2010]{Baccianella2010}
Baccianella, S., Esuli, A., and Sebastiani, F. (2010).
\newblock Sentiwordnet 3.0: An enhanced lexical resource for sentiment analysis
  and opinion mining.
\newblock In Chair), N. C.~C., Choukri, K., Maegaard, B., Mariani, J., Odijk,
  J., Piperidis, S., Rosner, M., and Tapias, D., editors, {\em Proceedings of
  the Seventh International Conference on Language Resources and Evaluation
  (LREC'10)}, Valletta, Malta. European Language Resources Association (ELRA).

\bibitem[Bautin et~al., 2008]{Bautin2008}
Bautin, M., Vijayarenu, L., and Skiena, S. (2008).
\newblock International sentiment analysis for news and blogs.
\newblock In {\em Proceedings of the International Conference on Weblogs and
  Social Media (ICWSM)}.

\bibitem[Bobicev et~al., 2010]{Bobicev2010}
Bobicev, V., Maxim, V., Prodan, T., Burciu, N., and Angheluş, V. (2010).
\newblock Emotions in words: Developing a multilingual wordnet-affect.
\newblock In Gelbukh, A., editor, {\em Computational Linguistics and
  Intelligent Text Processing}, volume 6008 of {\em Lecture Notes in Computer
  Science}, pages 375--384. Springer Berlin Heidelberg.

\bibitem[Brooke et~al., 2009]{Brooke2009}
Brooke, J., Tofiloski, M., and Taboada, M. (2009).
\newblock Cross-linguistic sentiment analysis: From english to spanish.
\newblock In {\em International Conference RANLP}, pages 50--54.

\bibitem[Cambria et~al., 2010]{Cambria2010}
Cambria, E., Speer, R., Havasi, C., and Hussain, A. (2010).
\newblock Sentic{N}et: A publicly available semantic resource for opinion
  mining.
\newblock {\em Artificial Intelligence}, pages 14--18.

\bibitem[Cettolo et~al., 2012]{Cettolo2012}
Cettolo, M., Girardi, C., and Federico, M. (2012).
\newblock Wit3: Web inventory of transcribed and translated talks.
\newblock In Cettolo, M., Federico, M., Specia, L., and Way, A., editors, {\em
  Proceedings of the 16th Annual Conference of the European Association for
  Machine Translation}, pages 261--268, Trento, Italy. European Association for
  Machine Translation.

\bibitem[Cohen, 1960]{Cohen1960}
Cohen, J. (1960).
\newblock {A Coefficient of Agreement for Nominal Scales}.
\newblock {\em Educational and Psychological Measurement}, 20(1):37.

\bibitem[Denecke, 2008]{Denecke2008}
Denecke, K. (2008).
\newblock {Using SentiWordNet for multilingual sentiment analysis}.
\newblock In {\em Data Engineering Workshop, 2008. ICDEW 2008. IEEE 24th
  International Conference on}, pages 507--512.

\bibitem[Diehn, 2013]{dw}
Diehn, S. (2013).
\newblock {Social media use evolving in Egypt}.
\newblock \url{http://www.dw.de/social-media-use-evolving-in-egypt/a-16930251},
  [Online; accessed 21-August-2014].

\bibitem[Ekman, 1992]{Ekman1992}
Ekman, P. (1992).
\newblock An argument for basic emotions.
\newblock {\em Cognition \& Emotion}, 6(3-4):169--200.

\bibitem[Fellbaum, 1998]{Miller1998}
Fellbaum, C. (1998).
\newblock {\em WordNet: An electronic lexical database}.
\newblock MIT press.

\bibitem[Fleiss et~al., 2013]{Fleiss2013}
Fleiss, J.~L., Levin, B., and Paik, M.~C. (2013).
\newblock {\em Statistical methods for rates and proportions}.
\newblock John Wiley \& Sons.

\bibitem[Fort, 2011]{Fort2011}
Fort, K. (2011).
\newblock {Corpus Linguistics : Inter-Annotator Agreements}.

\bibitem[Ghorbel, 2012]{Ghorbel2012}
Ghorbel, H. (2012).
\newblock Experiments in cross-lingual sentiment analysis in discussion forums.
\newblock In Aberer, K., Flache, A., Jager, W., Liu, L., Tang, J., and Guéret,
  C., editors, {\em Social Informatics}, volume 7710 of {\em Lecture Notes in
  Computer Science}, pages 138--151. Springer Berlin Heidelberg.

\bibitem[Green, 1997]{Green1997}
Green, A.~M. (1997).
\newblock Kappa statistics for multiple raters using categorical
  classifications.
\newblock In {\em Proceedings of the Twenty-Second Annual Conference of SAS
  Users Group}, San Diego, USA.

\bibitem[Gutierrez et~al., 2016]{gutierrez2016detecting}
Gutierrez, E., Shutova, E., Lichtenstein, P., de~Melo, G., and Gilardi, L.
  (2016).
\newblock Detecting cross-cultural differences using a multilingual topic
  model.
\newblock {\em Transactions of the Association for Computational Linguistics},
  4:47--60.

\bibitem[Hajmohammadi et~al., 2015]{hajmohammadi2015combination}
Hajmohammadi, M.~S., Ibrahim, R., Selamat, A., and Fujita, H. (2015).
\newblock Combination of active learning and self-training for cross-lingual
  sentiment classification with density analysis of unlabelled samples.
\newblock {\em Information Sciences}, 317:67--77.

\bibitem[Krippendorff, 1980]{Krippendorff1980}
Krippendorff, K. (1980).
\newblock {\em Content Analysis: An Introduction to Methodology}.
\newblock Sage Publications, Inc.

\bibitem[Landis and Koch, 1977]{Landis1977}
Landis, R. and Koch, G. (1977).
\newblock The measurement of observer agreement for categorical data.
\newblock {\em Biometrics}, 33(1):159--174.

\bibitem[Larkey et~al., 2007]{Larkey2007Light}
Larkey, L., Ballesteros, L., and Connell, M. (2007).
\newblock Light stemming for arabic information retrieval.
\newblock In {\em Arabic computational morphology}, pages 221--243. Springer.

\bibitem[Lloyd, 2006]{Lloyd2006}
Lloyd, L. (2006).
\newblock {\em Lydia: a system for the large scale analysis of natural language
  text}.
\newblock State University of New York at Stony Brook.

\bibitem[Ma and Zakhary, 2009]{Dalal2009}
Ma, X. and Zakhary, D. (2009).
\newblock Arabic newswire english translation collection.
\newblock {\em Linguistic Data Consortium, Philadelphia}.

\bibitem[NIST, 2010]{NIST}
NIST, M. (2010).
\newblock Nist 2008/2009 open machine translation (openmt) evaluation.
\newblock {\em Linguistic Data Consortium, Philadelphia}.

\bibitem[Pang and Lee, 2004]{Pang2004}
Pang, B. and Lee, L. (2004).
\newblock A sentimental education: Sentiment analysis using subjectivity
  summarization based on minimum cuts.
\newblock In {\em Proceedings of the 42nd Annual Meeting on Association for
  Computational Linguistics}, page 271. Association for Computational
  Linguistics.

\bibitem[Pang and Lee, 2005]{Pang2005}
Pang, B. and Lee, L. (2005).
\newblock Seeing stars: exploiting class relationships for sentiment
  categorization with respect to rating scales.
\newblock In {\em Proceedings of the 43rd Annual Meeting on Association for
  Computational Linguistics}, ACL '05, pages 115--124, Stroudsburg, PA, USA.
  Association for Computational Linguistics.

\bibitem[Pang and Lee, 2008]{Pang2008}
Pang, B. and Lee, L. (2008).
\newblock Opinion mining and sentiment analysis.
\newblock {\em Found. Trends Inf. Retr.}, 2(1-2):1--135.

\bibitem[Rafalovitch and Dale, 2009]{Rafalovitch2009}
Rafalovitch, A. and Dale, R. (2009).
\newblock United nations general assembly resolutions: A six-language parallel
  corpus.
\newblock In {\em Proceedings of Machine Translation Summit XII: Posters}.

\bibitem[Saad, 2015]{Saad2015phd}
Saad, M. (2015).
\newblock {\em {Mining Documents and Sentiments in Cross-lingual Context}}.
\newblock PhD thesis, Université de Lorraine.

\bibitem[Saad and Ashour, 2010]{Saad2010morphological}
Saad, M. and Ashour, W. (2010).
\newblock {Arabic Morphological Tools for Text Mining}.
\newblock In {\em EEECS’10 the 6th International Symposium on Electrical and
  Electronics Engineering and Computer Science}, pages 112--117. European
  University of Lefke, Cyprus.

\bibitem[Saad et~al., 2013]{Saad2013cilc}
Saad, M., Langlois, D., and Smaili, K. (2013).
\newblock {Extracting Comparable Articles from Wikipedia and Measuring their
  Comparabilities}.
\newblock {\em Procedia - Social and Behavioral Sciences}, 95(0):40 -- 47.
\newblock Corpus Resources for Descriptive and Applied Studies. Current
  Challenges and Future Directions: Selected Papers from the 5th International
  Conference on Corpus Linguistics (CILC2013).

\bibitem[Saad et~al., 2014]{Saad2014lrec}
Saad, M., Langlois, D., and Smaili, K. (2014).
\newblock {Building and Modelling Multilingual Subjective Corpora}.
\newblock In {\em Proceedings of the Ninth International Conference on Language
  Resources and Evaluation (LREC'14)}, Reykjavik, Iceland. European Language
  Resources Association (ELRA).

\bibitem[Severyn et~al., 2016]{severyn2016multi}
Severyn, A., Moschitti, A., Uryupina, O., Plank, B., and Filippova, K. (2016).
\newblock Multi-lingual opinion mining on youtube.
\newblock {\em Information Processing and Management}, 52(1):46--60.
\newblock Emotion and Sentiment in Social and Expressive Media.

\bibitem[Strapparava and Mihalcea, 2007]{Strapparava2007}
Strapparava, C. and Mihalcea, R. (2007).
\newblock Semeval-2007 task 14: affective text.
\newblock In {\em Proceedings of the 4th International Workshop on Semantic
  Evaluations}, SemEval '07, pages 70--74, Stroudsburg, PA, USA. Association
  for Computational Linguistics.

\bibitem[Torii et~al., 2011]{Torii2011}
Torii, Y., Das, D., Bandyopadhyay, S., and Okumura, M. (2011).
\newblock Developing japanese wordnet affect for analyzing emotions.
\newblock In {\em Proceedings of the 2nd Workshop on Computational Approaches
  to Subjectivity and Sentiment Analysis}, pages 80--86. Association for
  Computational Linguistics.

\end{thebibliography}

\end{document}